\documentclass{llncs}

\input glyphtounicode\pdfgentounicode=1
\usepackage{amsmath}
\usepackage{amssymb}
\usepackage{graphicx}
\usepackage[FIGTOPCAP]{subfigure}
\usepackage{wrapfig}
\usepackage{soul}

\usepackage[T1]{fontenc}
\usepackage{capt-of}
\usepackage[font=small,labelfont=bf,tableposition=top]{caption}
\DeclareCaptionLabelFormat{andtable}{#1~#2  \&  \tablename~\thetable}

\usepackage{floatrow}
\usepackage[T1]{fontenc}
\usepackage[utf8]{inputenc}
\usepackage[font=small,labelfont=bf]{caption}
\newfloatcommand{capbtabbox}{table}[][\FBwidth]
\usepackage[table]{xcolor}

\usepackage{graphics}

\usepackage[amssymb]{SIunits}
\usepackage{stackengine}
\def\delequal{\mathrel{\stackon[1pt]{=}{$\scriptscriptstyle\Delta$}}}
\begin{document}
\title {Bayesian Image Quality Transfer with CNNs: Exploring Uncertainty in dMRI Super-Resolution}
%


\author{Ryutaro Tanno\inst{1,3}\and Daniel Worrall\inst{1}\and Aurobrata Ghosh\inst{1}\and Enrico Kaden\inst{1}\and Stamatios N. Sotiropoulos\inst{2}\and Antonio Criminisi\inst{3} \and Daniel C. Alexander\inst{1}}
\institute{Department of Computer Science, University College London, UK,\\
\and FMRIB Centre, University of Oxford, UK,\\
\and Microsoft Research Cambridge, UK}
\maketitle    
\begin{abstract}
        
       In this work, we investigate the value of uncertainty modelling 
       in 3D super-resolution with convolutional neural networks (CNNs). Deep learning has shown success in a plethora of medical image transformation problems, such 
       as super-resolution (SR) and image synthesis. However, the highly ill-posed nature 
       of such problems results in inevitable ambiguity in the learning of networks. 
       We propose to account for \textit{intrinsic uncertainty} through a per-patch 
       heteroscedastic noise model and for \textit{parameter uncertainty} through approximate Bayesian inference in the form of variational dropout. We show that the combined benefits of both lead to the state-of-the-art performance SR of diffusion MR brain images in terms of errors compared to ground truth. We further show that the reduced error scores produce tangible benefits in downstream tractography. In addition, the probabilistic nature of the methods naturally confers a mechanism to quantify uncertainty over the super-resolved output. We demonstrate 
       through experiments on both healthy and pathological brains the potential utility 
       of such an uncertainty measure in the risk assessment of the super-resolved images 
       for subsequent clinical use.
		
\end{abstract}

\section{Introduction and Background}

Algorithmic and hardware advancements of non-invasive imaging techniques, such as MRI, continue to push the envelope of quality and diversity of obtainable information of the underlying anatomy. However, their prohibitive cost and lengthy acquisition time often hinder the translation of such technological innovations into clinical practice. Poor image quality limits the accuracy of subsequent analysis, potentially leading to false clinical conclusions. Therefore, methods which can efficiently and reliably boost scan quality are in demand. 

Numerous machine learning based methods have been proposed for various forms of image enhancement, generally via supervised regression of low quality (e.g., clinical) against high quality (e.g., experimental) image content. Alexander et al. \cite{alexander2014image} propose a general framework for supervised image quality enhancement, which they call image quality transfer (IQT). They demonstrated this with a random forest (RF) implementation of super-resolution (SR) of brain diffusion tensor images (DTIs) and estimation of advanced microstructure parameter maps from sparse measurements. More recently, deep learning has shown additional promise in this kind of task. For example, \cite{oktay2016multi} proposed a CNN model to upsample a stack of 2D MRI cardiac volumes in the through-plane direction. The SR mapping is learnt from 3D cardiac volumes of nearly isotropic voxels, requiring a clinically impractical sequence due to its long scan time. Another application of CNNs is the prediction of 7T images from 3T MRI \cite{bahrami2016convolutional}, where both contrast and resolution are enhanced. Current methods typically commit to a single prediction, leaving users with no measure of prediction reliability. One exception is Bayesian IQT \cite{tanno2016bayesian}, which uses a piece-wise linear Bayesian model, implemented with a RF to quantify predictive uncertainty over high-resolution (HR) DTIs and demonstrate its utility as a surrogate measure of accuracy. 

This paper proposes a new implementation of Bayesian IQT via CNNs. This involves two key innovations in CNN-based models: 1) we extend the subpixel CNN of \cite{shi2016real}, previously limited to 2D images, to 3D volumes, outperforming previous models in accuracy and speed on a DTI SR task; 2) we devise new architectures enabling estimates of different components of the uncertainty in the SR mapping. The first enables us to bring the performance benefits of deep learning to this important problem, as well as reducing computation time to super-resolve the entire brain DTI in 1 s. For our second contribution, we describe two kinds of uncertainty which arise when tackling image enhancement problems. The first kind of uncertainty, which we call \emph{intrinsic uncertainty} is defined as the irreducible variance of the statistical mapping from low-resolution (LR) to HR. This inherent ambiguity arises from the fact that the LR to HR problem is one-to-many, and is present independent of the amount of data we collect. We model the variation in intrinsic uncertainty over different structures within the anatomy through a per-patch heteroscedastic noise model \cite{nix1994estimating}. The second kind of uncertainty, which we call \emph{parameter uncertainty}, quantifies the degree of ambiguity in the model parameters that best explain the observed data, which arises from the finite training set. We account for it through approximate Bayesian inference in the form of variational dropout \cite{kingma2015variational}. 

We first evaluate the performance of the proposed CNN methods and the benefits of uncertainty modelling by measuring the deviation from the ground truth on standard metrics. Human Connectome Project (HCP) dataset \cite{sotiropoulos2013advances} and the Lifespan dataset (\url{http://lifespan.humanconnectome.org/}) are used for the quantitative comparison. We also test the benefits of the CNN-based methods in downstream tractography through SR of Mean Apparent Propagator (MAP)-MRI \cite{ozarslan2013mean}. Lastly, we investigate the utility of uncertainty maps over the predicted HR image from the probabilistic CNN methods by testing on images of both healthy subjects and brain tumour patients.

\section{Method}
As in \cite{alexander2014image,oktay2016multi,bahrami2016convolutional}, we formulate the SR task as a patch-wise regression where an input LR image is split into smaller overlapping sub-volumes and the resolution of each is sequentially enhanced. For this, we propose a baseline CNN, on which we build by introducing two complementary ways of accounting for uncertainty to achieve a more robust model. 
\\
\textbf{Baseline network:} Efficient subpixel-shifted convolutional network (ESPCN) \cite{shi2016real} is a recently proposed method with the capacity to perform real-time per-frame SR of videos while retaining cutting-edge performance.  We extend this method to 3D and use this as our baseline model (3D-ESPCN). Most CNN-based SR techniques \cite{oktay2016multi,dong2016image,johnson2016perceptual} first up-sample a low-resolution (LR) input image (e.g. through bilinear interpolation, deconvolution, fractional-strided convolution, etc.) and then refine the high-resolution (HR) estimate through a series of convolutions. These methods suffer from the fact that $(1)$ the up-sampling can be a lossy process and $(2)$ refinement in the HR-space has a higher computational cost than in the LR-space. ESPCN performs convolutions in the LR-space, upsampling afterwards. The reduced resolution of feature maps dramatically decreases the computational and memory costs, which is more pronounced in 3D. 

\begin{figure}[t]
	\includegraphics[width=10.0cm]{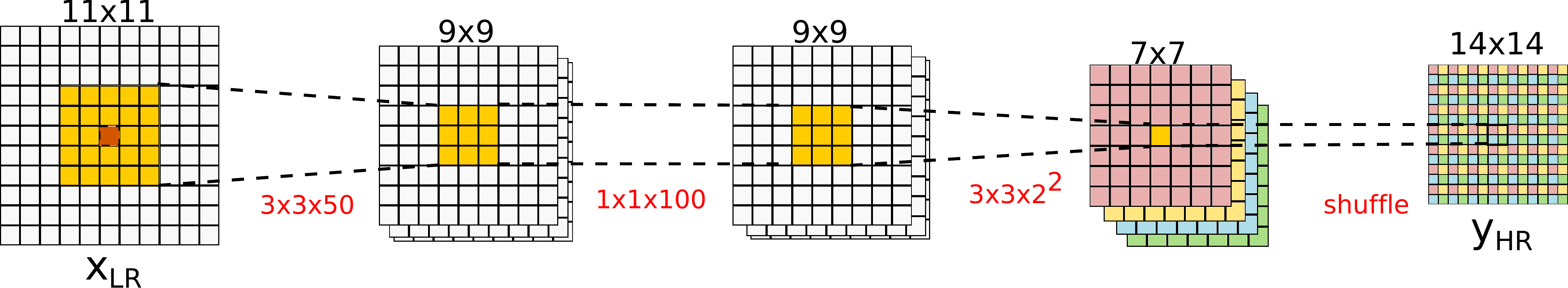}
	\centering	
	\small 
	\caption{2D illustration of the baseline network with upsampling rate, $r=2$. The receptive field of the central $2^2$ output activations is shown in yellow.} 
	\label{fig:2DESPCN}
	\vspace{-10pt}
\end{figure}

More specifically the ESPCN is a fully convolutional network, with a special \emph{shuffling operation} on the output (see Fig. \ref{fig:2DESPCN}). The fully convolutional part of the network consists of 3 convolutional layers, each followed by a ReLU, where the final layer has $cr^2$ channels, $r$ being the upsampling rate. The shuffling operation takes an input of shape $h\times w\times cr^2$ and remaps pixels from different channels into different spatial locations in the HR output, producing a $rh\times rw\times c$ image, where $h$, $w$ and $c$ denote height, width and number of channels. The 3D version of this shufflling operation $\mathcal{S}$ is mathematically given by $\mathcal{S}(F)_{i,j,k,c} =  F_{[i/r],[j/r],[k/r],(r^3-1)c + \text{mod}(i,r) + r\cdot \text{mod}(j,r) + r^3\cdot \text{mod}(k,r)}$ where $F$ is the pre-shuffled feature maps. \cite{shi2016real} showed that the combined effects of the last convolution and shuffling is effectively a learned interpolation. 

At test time, the network takes each subvolume $\mathbf{x}$ in a LR image, and predicts the corresponding HR sub-volume $\mathbf{y}$. The network increases the resolution of the central voxel of each receptive field, e.g. the central $2^3$ output voxels are estimated from the corresponding $5^3$ receptive field in the input, coloured yellow in Fig. \ref{fig:2DESPCN}. By tessellating the predictions from shifted inputs $\mathbf{x}$, the whole HR volume is reconstructed. 

\begin{figure}
	\vspace{-20pt}
	\includegraphics[width=10cm]{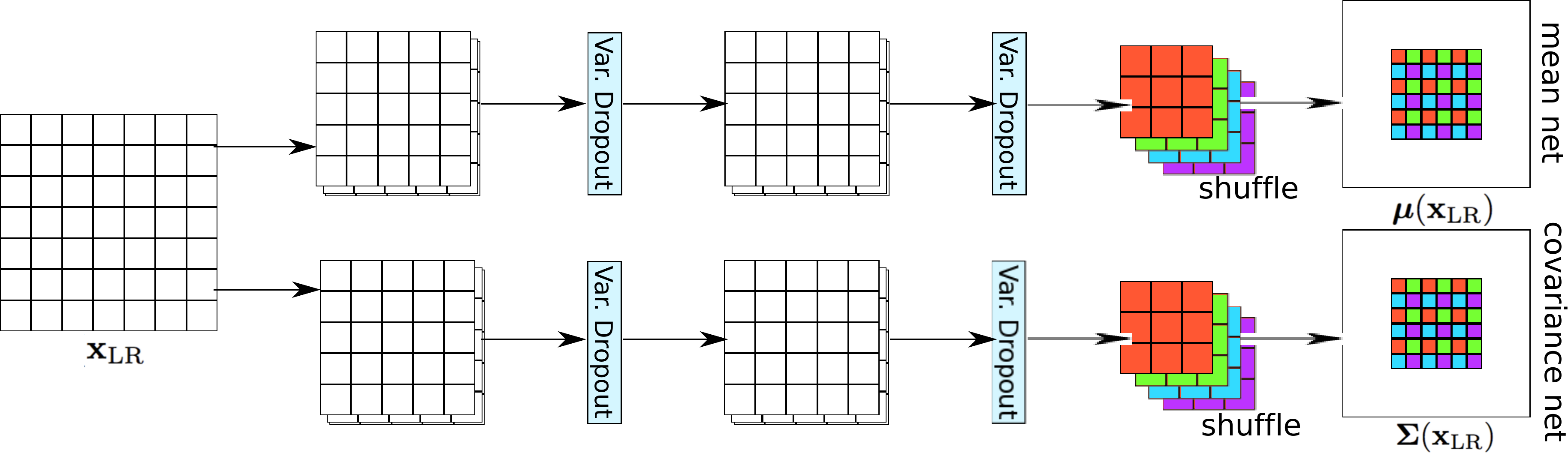}
	\centering	
	\small 
	\caption{2D illustration of a heteroscedastic network with variational dropout. Diagonal covariance is assumed. The top 3D-ESPCN estimates the mean and the bottom one estimates the covariance matrix of the likelihood. Variational dropout is applied to feature maps after every convolution.} 
	\label{fig:heterovar}
	\vspace{-7pt}
\end{figure}

Given a training set $\mathcal{D}=\{(\mathbf{x}_i,\mathbf{y}_i)\}_{i=1}^N$, we optimize the network parameters by minimising the sum of per-pixel mean-squared-error (MSE) between the ground truth $\mathbf{y}$ and the predicted HR patch $\mu_{\theta}(\mathbf{x})$ over the training set. $\theta$ denotes all network parameters. This is equivalent to minimising the negative log likelihood (NLL) under the Gaussian noise model $p(\mathbf{y}|\mathbf{x},\mathbf{\theta}) = \mathcal{N}(\mathbf{y}; \mu_{\theta}(\mathbf{x}), \sigma^2I)$.  Here, HR patches are modelled as a deterministic function of LR patches corrupted by isotropic noise with variance $\sigma^2$. On the assumption that the model is correct, the variance $\sigma^2$ signifies the degree of irreducible uncertainty in the prediction of $\mathbf{y}$ given $\mathbf{x}$, and thus the \textit{intrinsic uncertainty} in the SR mapping defined in the introduction. However, the quality of this intrinsic uncertainty estimation is limited by the quality of likelihood model; the baseline network assumes constant uncertainty across all spatial locations and image channels, which is over-simplistic for most medical images.
\\
\textbf{Heteroscedastic likelihood:\label{sec:hetero}} Here we introduce a \textit{heteroscedastic} noise model to approximate the variation in intrinsic uncertainty across the image. The likelihood becomes $p(\mathbf{y}|\mathbf{x},\mathbf{\theta}_1, \mathbf{\theta}_2) = \mathcal{N}(\mathbf{y}; \mu_{\theta_1}(\mathbf{x}),\Sigma_{\theta_2}(\mathbf{x}))$ where both mean and covariance are estimated by two separate 3D-ESPCNs $\mu_{\theta_1}(\cdot)$ and $\Sigma_{\theta_2}(\cdot)$ as a function of the input. The mean network makes predictions and the covariance network estimates the intrinsic uncertainty (see Fig. \ref{fig:heterovar}). We only use the diagonal of $\Sigma_{\theta_2}(\mathbf{x})$, which quantifies estimated intrinsic uncertainty over individual components in $\mu_{\theta_1}(\mathbf{x})$. 

Adapting the NLL to heteroscedastic model gives 
$\mathcal{L}_{\theta}(\mathcal{D}) = \mathcal{M}_{\theta}(\mathcal{D}) + \mathcal{H}_{\theta}(\mathcal{D})$
with $\mathcal{M}_{\theta}(\mathcal{D}) = \frac{1}{N}\sum_{i=1}^{N}\|\mathbf{y}_i-\mu_{\theta_1}(\mathbf{x}_i)\|_{\Sigma^{-1}_{\theta_2}(\mathbf{x}_i)}^2$ i.e. mean squared Mahalanobis distance and $\mathcal{H}_{\theta}(\mathcal{D}) = \frac{1}{N}\sum_{i=1}^{N}\text{log det}\Sigma_{\theta_2}( \mathbf{x}_i)$ i.e. mean differential entropy. Intuitively, $\mathcal{M}_{\theta}(\mathcal{D})$ seeks to minimise the weighted MSE under the covariance while $\mathcal{H}_{\theta}(\mathcal{D})$ keeps the `spread' of $\Sigma_{\theta_2}(\mathbf{x})$ under control. 
%
%
\\
\textbf{Bayesian inference through variational dropout:} The baseline 3D-ESPCN and heteroscedastic model neglect \textit{parameter uncertainty}, relying on a single estimate of the network parameters. In medical imaging where data size is commonly limited, this point-estimate approach potentially leads to overfitting. We combat this with a Bayesian approach, averaging over all possible models $p(\mathbf{y}|\mathbf{x}, \theta)$ weighted by the (posterior) probability of the parameters given the training data, $p(\theta|\mathcal{D})$. Mathematically this is $p(\mathbf{y}|\mathbf{x}, \mathcal{D}) = \mathbb{E}_{p(\theta|\mathcal{D})}[p(\mathbf{y}|\mathbf{x}, \theta)]$. However, this expectation is intractable because: 1) evaluation of $p(\theta|\mathcal{D})$ is intractable, and 2) $p(\mathbf{y}|\mathbf{x}, \theta)$ is too complicated for the expectation to be computed in closed form.  Variational dropout \cite{kingma2015variational} addresses this problem for neural networks, using a form of variational inference where the posterior $p(\theta|\mathcal{D})$ is approximated by a factored Gaussian distribution $q_{\phi}(\theta) = \prod_{ij} \mathcal{N}(\theta_{ij}; m_{ij}, s_{ij}^2)$. The algorithm injects Gaussian noise into the weights during training, where the amount of noise is controlled by $\phi=\{m_{ij}, s_{ij}^2\}$, which are learnt.

At test time, given a LR input $\mathbf{x}$, we estimate the mean and covariance of the approximate predictive distribution $q_\phi^*(\mathbf{y}|\mathbf{x}) \delequal \mathbb{E}_{q_{\phi}(\theta)}[p(\mathbf{y}|\mathbf{x},\theta)]$ with the MC estimators $\hat{\mu}_{\mathbf{y}|\mathbf{x}} \delequal \frac{1}{T}\sum_{t=1}^T\mu_{\theta_{1}^{t}}(\mathbf{x})$ and $\hat{\Sigma}_{\mathbf{y}|\mathbf{x}} \delequal \frac{1}{T} \sum_{t=1}^T\Big{(}\Sigma_{\theta_{2}^{t}}(\mathbf{x})+\mu_{\theta_1^t}(\mathbf{x})\mu_{\theta_1^t}(\mathbf{x})^{T}\Big{)}-\hat{\mu}_{\mathbf{y}|\mathbf{x}} \hat{\mu}_{\mathbf{y}|\mathbf{x}}^T$, where $\theta^t=(\theta^t_1,\theta^t_2)$ are samples of the network parameters (i.e. the filters) from the approximate posterior $ q_{\phi}(\theta)$. We use the sample mean as the final prediction of an HR patch and the diagonal of the sample variance as the corresponding uncertainty. When we use the baseline model, the first term in the sample variance reduces to $\sigma^2I$.
\\
\textbf{Implementation details:} We employed a common protocol for the training of all networks. We minimized the loss using ADAM \cite{kingma2014adam} for $200$ epochs with learning rate $10^{-3}$. The best performing model was selected on a validation set. 

As in \cite{shi2016real}, we use a minimal architecture for the baseline 3D-ESPCN, consisting of $3$ convolutional layers with filters $(3,3,3,50)\to(1,1,1,100)\to(3,3,3,r^3c)$ where $r$ is upsampling rate and $c$ is the number of channels. The filter sizes are chosen so a $(5,5,5)$ LR patch maps to a $(r,r,r)$ HR patch, which mirrors competing random forest based methods \cite{alexander2014image,tanno2016bayesian} for a fair comparison. The heteroscedastic network of Section \ref{sec:hetero} is formed of two 3D-ESPCNs, separately estimating the mean and standard deviations. Positivity of the standard deviations is enforced by passing the output through a \emph{softplus} function. For variational dropout we tried two flavours: Var.(I) optimises per-weight dropout rates, and Var.(II) optimises per-filter dropout rates. Variational dropout is applied to both the baseline and heteroscedastic models without changing the architectures. 

All models are trained on datasets generated from 8 randomly selected HCP subjects \cite{sotiropoulos2013advances}, each consisting of $90$ diffusion weighted images (DWIs) of voxel size $1.25^3\text{ mm}^3$ with $b = 1000 \text{ s/mm}^2$. The training set is created by sampling HR subvolumes inside the brain region from the ground truth DTIs (or MAP-MRI coefficients) and then artificially downsampling to generate the LR counterparts. Downsampling is done in the raw DWI by a factor of $r$ by taking a block-wise mean and then the DT or MAP coefficients are subsequently computed. Each network is trained on $\sim 4000$ pairs of input/output patches of size $11^3c$ and $(7r)^3c$, amounting to $\sim1.4\times10^6$ receptive field patch pairs of dimensions $5^3c$ and $r^3c$, which is roughly the same size as the maximal training set used in RF-IQT \cite{alexander2014image}. It takes under $30/120$ mins to train a single network on DTI/MAP-MRI data on 1 TITAN X GPU.
\section{Experiments and Results}
\textbf{Performance comparison for DTI SR:}
We evaluate the performance of our models for DTI SR on two HCP datasets. The first contains 8 unseen subjects from the same HCP cohort used for training. The second consists of $10$ subjects from the HCP Lifespan dataset. The latter tests generalisability, as they are acquired with different protocols, at lower resolution (1.5 mm isotropic), and on subjects of a different age range (45-75) to the original HCP data (22-36). We perform $\times 2$ upsampling in each direction, measuring reconstruction accuracy with RMSE, PSNR and MSSIM on the interior and the exterior separately as shown in Fig. 3(b). This is important, as the estimation problem is quite different in boundary regions, but remains valuable particularly for applications like tractography where seed or target regions are often in the cortical surface of the brain.  We only present the RMSE results, but the derived conclusions remain the same for the other two metrics. 

Fig. 3(a) shows our baseline achieves $8.5\%/39.8\%$ reduction in RMSE on the HCP dataset on the interior/exterior regions with respect to the best published method, BIQT-RF\cite{tanno2016bayesian}. Note that IQT-RF and BIQT-RF are only trained on interior patches, and SR on boundary patches requires a separate ad-hoc procedure. Despite including exterior patches in training our model, which complicates the learning task, the baseline CNN out-performs the RF methods on both regions---this goes for the Lifespan dataset too. The 3D-ESPCN estimates whole HR volumes <10 s on a CPU and $\sim1$ s on a GPU, while BIQT-RF takes $\sim10$ mins with 8 trees. Faster reconstruction is achieved using input patches of size $22^3$, twice as large as the training input size $11^3$. 
\begin{figure}[t]
	\vspace{- 15pt}
	\centering
	\subfigure[Performance comparison]{\includegraphics[width=10.0cm]{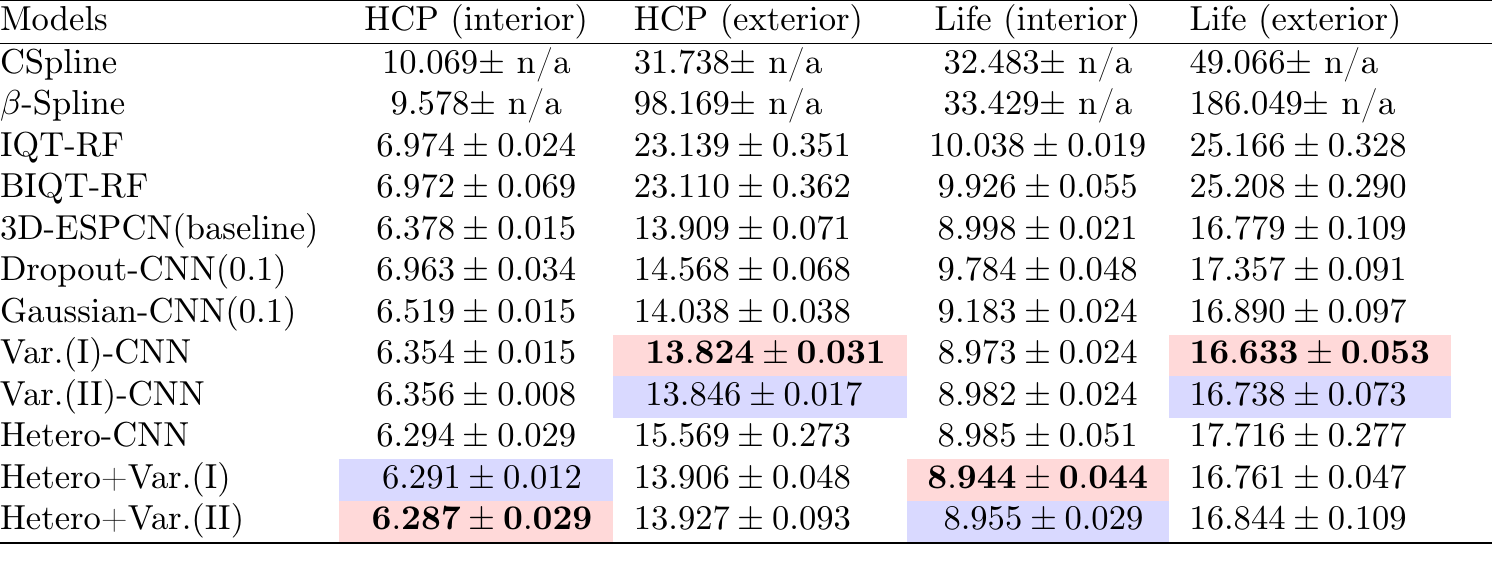}\label{fig:performance_compare}}
	\subfigure[Mask]{\includegraphics[width=1.75cm]{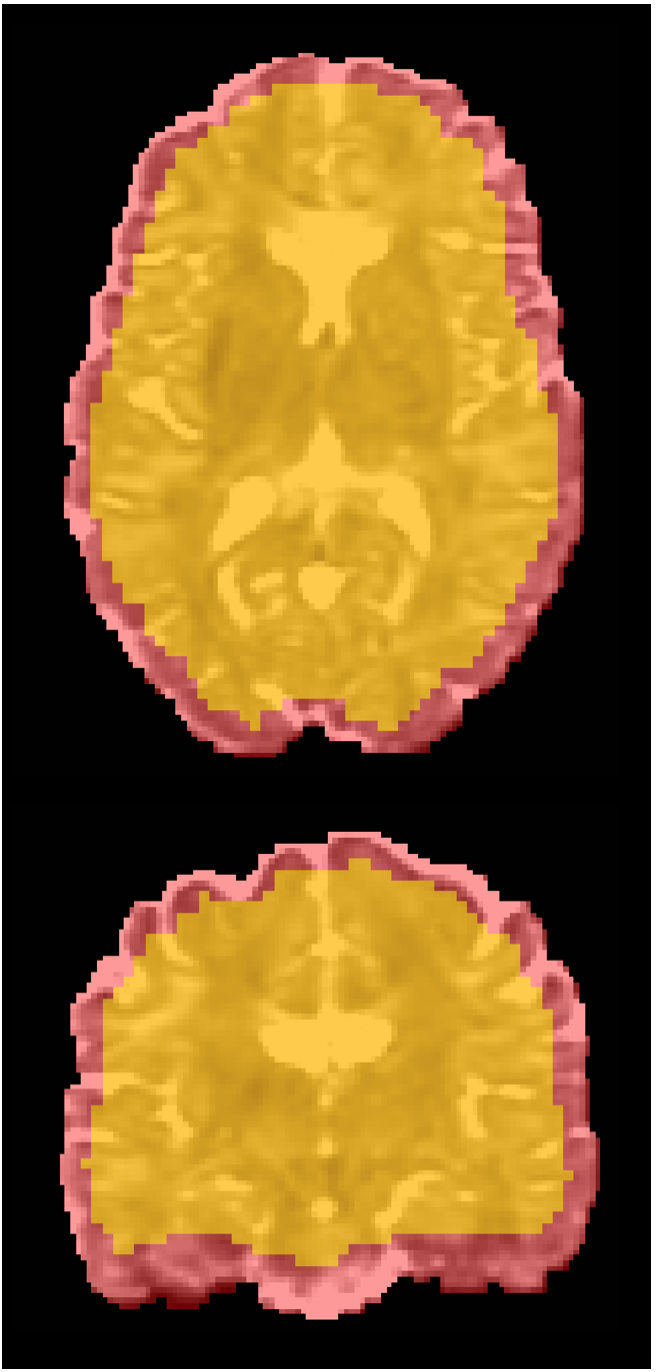}\label{fig:mask}}
	\small
	\caption{(a) RMSE on HCP and Lifespan dataset for different upsampling methods. For each method, an ensemble of $10$ models are trained on different training sets, and the mean/std of the average errors over $8$ test subjects are computed over the ensemble. Best results in bold red, and the second best in blue. (b) Interior (yellow) and exterior region (red).} 
	\vspace{- 15pt}
\end{figure}

Heteroscedastic network further improves on the performance of 3D-ESPCN with high statistical significance on the interior region for both HCP and Lifespan data ($p<10^{-3}$). However, poorer performance is observed on the exterior than the baseline. Using $200$ weight samples, we see Var.(I)-CNN performs best on both datasets on the exterior region. Combination of heteroscedastic model and variational dropout (i.e. Hetero+Var.(I) or (II)) leads to the top 2 performance on both datasets on the interior region and reduces errors on the exterior to the level comparable or better than the baseline. 

The performance difference of heteroscedastic network between the interior and the exterior region roots from the loss function. The term $\mathcal{M}_{\theta}(\mathcal{D})$ imposes a larger penalty on the regions with smaller intrinsic uncertainty. The network therefore allocates more of its resources towards the lower noise regions where the statistical mapping from the LR to HR space is less ambiguous. The dramatic reduction on the exterior error from variational dropout indicates its regularisation effect against such overfitting, and as a result also improves the robustness of prediction on the interior.
\\
\textbf{Tractography with MAP-MRI SR:} Reconstruction accuracy does not reflect real world utility. We thus further assessed SR quality with a tractography experiment on the Prisma dataset, which contains two DWIs of the same subject from a Siemens Prisma 3T scanner, with 1.35 mm and 2.5 mm resolution. The b-values and gradient directions match the HCP protocol. An ensemble of $8$ hetero+var.(I) CNNs super-resolves the MAP-MRI coefficients \cite{ozarslan2013mean} derived from the LR DWIs (2.5 mm), then the HR MAP volume is used to predict the HR DWIs (1.25 mm). The final prediction is computed as the average estimate weighted by the inverse covariance as in RF-IQT. We also generate HR datasets by using IQT-RF and linear interpolation.
\begin{figure}[t]
\vspace{-4em}
	\centering
	\includegraphics[width=\linewidth]{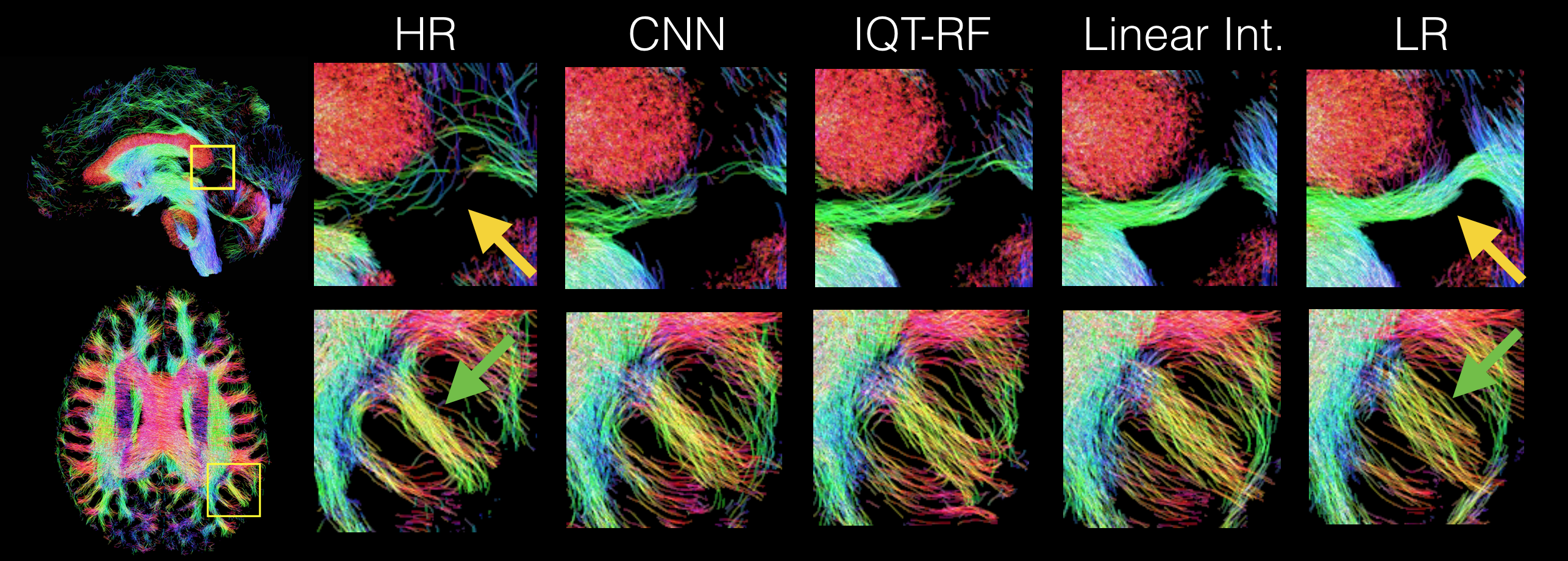}\label{fig:uncertainty_map}
	\small
	\caption{Tractography on Prisma dataset for different methods. From left to right: (i) HR acquisition, (ii) CNN prediction; (iii) RF; (iv) Linear interpolation; (v) LR acquisition.} 
	\label{fig:tract}
	\vspace{-1em}
\end{figure}

Fig. \ref{fig:tract} shows streamline maps of the probabilistic tractography \cite{tournier2010improved} for the original LR/HR data and various upsampled images, and focuses on examples that highlight the benefits of reduced SR reconstruction errors. In the top row, tractography on the LR data produces a false-positive tract under the corpus callosum (yellow arrow in the 1st row), which tractography at HR avoids. Reconstructured HR images from IQT-RF and CNN avoid the false positive better than linear intepolation. Note that we do not expect to reproduce the HR tractography map exactly, as the HR and LR images do not aligned exactly. The bottom row shows shaper recovery of small gyral white matter pathways (green arrow) at HR than LR resulting from reduced partial volume effect. CNN reconstruction produces a sharper pathway than RF-IQT and linear interpolation.
\\
\textbf{Visualisation of predictive uncertainty:} We measure the expectation and variance of \textit{mean diffusivity} (MD) and \textit{fractional anisotropy} (FA) with respect to the predictive distribution $q_\phi^*(\mathbf{y}|\mathbf{x})$ of Hetero+Var. (I) by MC sampling. Comparative results are shown in Fig. 5(a), where we drew $200$ samples of HR DTIs from the predictive distribution of the Hetero+Var.(I). The uncertainty map is highly correlated with the error maps. In particular, the MD uncertainty map captures subtle variations within the white matter and central CSF, which demonstrates the potential utility of the uncertainty map as a surrogate measure of accuracy.
\begin{figure}[ht]
\vspace{-2em}
	\centering
	\subfigure[Uncertainty propagation]{\includegraphics[width=9.35cm]{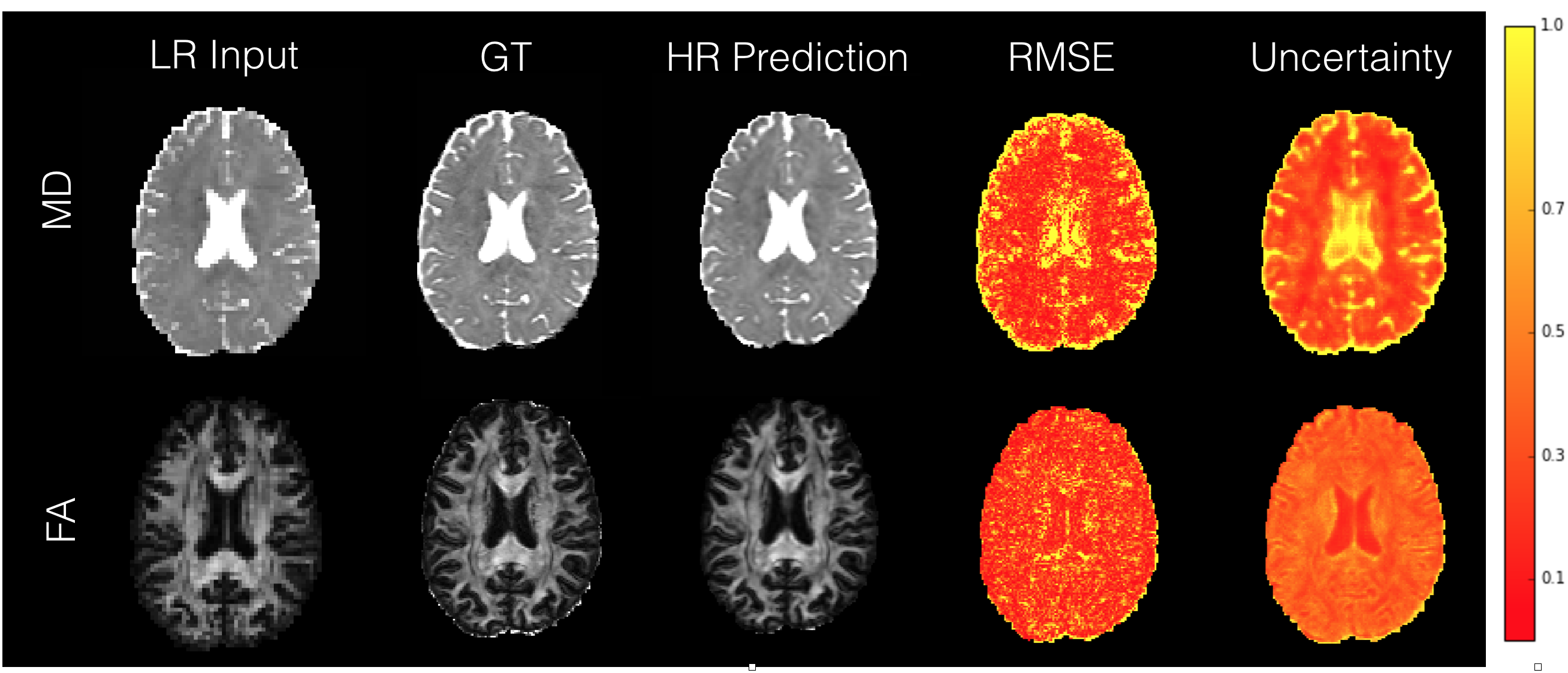}\label{fig:uncertainty_map}}
	\subfigure[Tumour]{\includegraphics[width=2.7cm]{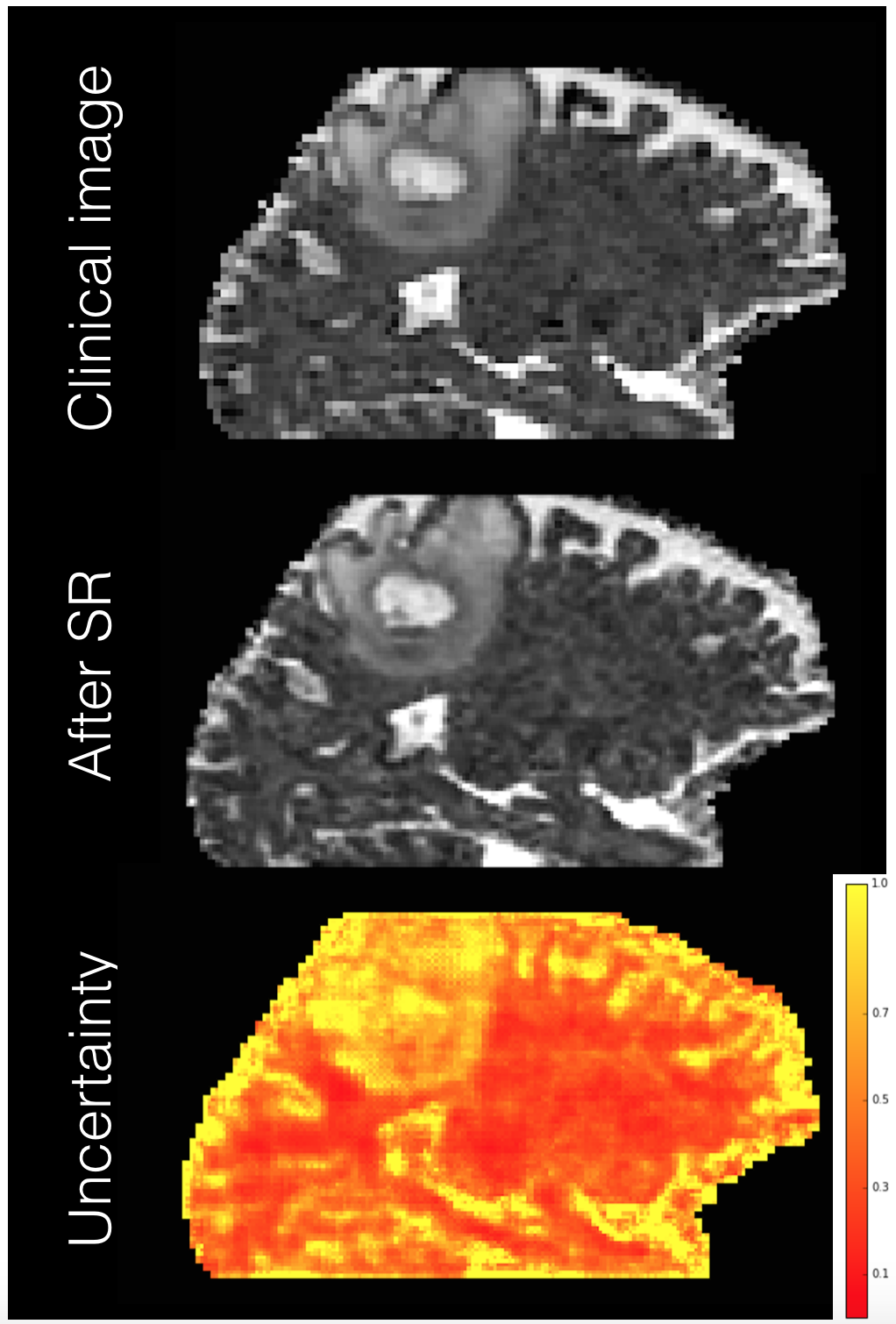}\label{fig:tumour}}
	\small
	\caption{(a) Comparison between RMSE and uncertainty maps for FA and MD computed on a HCP subject. LR input, ground truth and HR prediction are also shown. (b) DTI SR on a brain tumour patient. From top to bottom: (i) MD computed from the original DTI; (ii) the estimated HR version; (iii) uncertainty.} 
	\vspace{-1em}
\end{figure}

Fig. 5(b) shows the best-performing SR model (Hetero+Var (II)) trained on a healthy HCP cohort applied to the DTI of a brain tumour patient. The raw data (DWI) with $b=700 \text{ s/mm}^2$ is processed as before with input voxel size $2^3 \text{ mm}^3$. We show the input, SR image and uncertainty map. The ground truth is unavailable but the estimated image sharpens the input without introducing noticeable artifacts. The uncertainty map shows high uncertainty on the tumour, not represented in the training data, again illustrating the potential of the uncertainty maps to flag potential low accuracy areas.
\section{Discussion}
We present a super-resolution algorithm based on 3D subpixel-CNNs with state-of-the-art accuracy and reconstruction efficiency on diffusion MRI datasets. An application to the MAP-MRI coefficients indicates benefits to tractography in comparison with the previous methods. We also demonstrate that assimilation of \textit{intrinsic} and \textit{parameter} uncertainty in the model leads to best predictive performance. The uncertainty map highly correlates with reconstruction errors and is able to highlight pathologies. This can be used to gain insight into the `black-box' prediction from CNNs. Understanding the behaviours of these uncertainty measures in unfamiliar test environments (e.g. pathologies) and their relations to predictive performance is an important future work for designing a more generalisable method. The presented ideas extend to many other quality enhancement problems in medical image analysis and beyond.

\subsubsection*{Acknowledgements.}
%
%
This work was supported by Microsoft scholarship. Data were provided in part by the HCP, WU-Minn Consortium (PIs: David Van Essen and Kamil Ugurbil; 1U54MH091657) funded by NIH and Wash. U. The tumour data were acquired as part of a study lead by Alberto Bizzi, MD at his hospital in Milan, Italy.

%
%

\bibliographystyle{splncs}
\bibliography{reference}

\end{document}